\documentclass{article}

\usepackage{microtype}
\usepackage{graphicx}
\usepackage{subcaption}
\usepackage{booktabs} 

\usepackage{hyperref}

\usepackage[accepted]{icml2026}

\makeatletter
\renewcommand{\ICML@appearing}{\textit{Mechanistic Interpretability Workshop
at the $\mathit{43}^{rd}$ International Conference on Machine Learning,
Seoul, South Korea, 2026.
Copyright 2026 by the author(s).}}
\makeatother

\usepackage{amsmath}
\usepackage{amssymb}
\usepackage{mathtools}
\usepackage{amsthm}

\usepackage[capitalize,noabbrev]{cleveref}

\theoremstyle{plain}

\theoremstyle{definition}

\theoremstyle{remark}

\usepackage[textsize=tiny]{todonotes}

\icmltitlerunning{Before the Last Token}

\begin{document}

\twocolumn[
  \icmltitle{Before the Last Token: Diagnosing Final-Token Safety Probe Failures}



  \icmlsetsymbol{equal}{*}

  \begin{icmlauthorlist}
    \icmlauthor{Shravan Doda}{comp}
  \end{icmlauthorlist}

  \icmlaffiliation{comp}{Kipo AI}

  \icmlcorrespondingauthor{Shravan Doda}{shravan@kipo.ai}

  \icmlkeywords{AI safety, jailbreak diagnostics, safety probes, hidden-state trajectories, mechanistic interpretability, language model monitoring}

  \vskip 0.3in
]



\printAffiliationsAndNotice{}  

\begin{abstract}
Final-token safety probes monitor a single hidden state after prompt prefill, but jailbreak prompts can contain probe-visible unsafe evidence distributed across earlier user-token representations that is missed by this readout. We study this prefill-time failure mode using SafeSwitch-style probes trained only on clean harmful and benign prompts across three instruction-tuned LLMs. The probes achieve high recall on clean harmful prompts, but miss many jailbreaks and can produce false positives on safety-adjacent benign prompts. Subspace analyses suggest that missed jailbreaks differ from clean benign prompts along directions that are poorly captured by the probe's representational subspace, and increasing probe bottleneck width does not reliably resolve this mismatch. Token-level prefill analyses reveal that probe-visible unsafe evidence often appears earlier in the sequence but is not exposed at the final-token readout, while naive max-pooling over token positions overfires on safe prompts. A simple PCA-HMM trajectory model, trained only on the same clean split, recovers many final-token misses from user-content prefill trajectories without the catastrophic false-positive behavior of naive token pooling, motivating trajectory-aware hidden-state analyses as diagnostic complements to final-token probes.
\end{abstract}

\section{Introduction}

Probe-based safety systems such as SafeSwitch \citep{han2025safeswitchsteeringunsafellm} offer a lightweight way to monitor language models before they generate unsafe outputs. These systems train classifiers on internal activations and use the resulting signal to trigger downstream safety interventions. In the prefill setting, a common design is to make this decision from the hidden state at the final prompt token. This is efficient, but it makes safety monitoring depend on one learned readout of one contextual representation.

We study a failure mode of this final-token readout. Across three instruction-tuned models, we train SafeSwitch-style probes on clean harmful and benign prompts from the original SafeSwitch train split. We construct jailbreak evaluations by applying jailbreak templates from HarmBench \citep{mazeika2024harmbenchstandardizedevaluationframework} to held-out SorryBench harmful requests. These probes achieve high recall on clean harmful prompts, yet miss many jailbreaks. Jailbreak prompts are not content-free distribution shifts: they contain an otherwise harmful request embedded inside adversarial framing, role play, obfuscation, or instruction-following context. Thus, a final-token miss does not imply that harmful content is absent from the prompt; it means the trained probe does not detect the wrapped request from the final prompt-token representation.

Our results suggest that jailbreak wrappers shift the final-token representation away from the clean harmfulness contrast captured by the probe. This is not simply a capacity problem: increasing the probe bottleneck width does not reliably improve jailbreak detection. Geometrically, contrast directions for missed jailbreaks have little energy in the probe-visible subspace, while the visible caught-versus-missed jailbreak contrast aligns strongly with the clean harmful-versus-benign training direction. This is consistent with a shortcut-like readout: jailbreaks are caught when their final-token representation still projects onto the clean harmful contrast, and missed when the wrapper-induced shift moves them away from that readout.

We then inspect token-by-token hidden-state trajectories during prompt prefill. Missed jailbreaks often contain non-final positions that receive high probe scores, including positions near the embedded harmful request, even though the final-token score is low. Naive pooling over token positions is not a solution because safe prompts can also produce high intermediate scores. This motivates trajectory-aware diagnostics: simple PCA-HMM models fit only on the same original SafeSwitch train split recover most final-token-probe jailbreak misses across models. We use these models diagnostically, not as deployed jailbreak detectors, to show that final-token probe failures reflect where and how wrapper-induced representations expose safety-relevant information in hidden-state trajectories.

\section{Related Work}

\paragraph{Representation geometry of refusal and harmfulness.}
Prior work has studied internal directions and subspaces associated with refusal, harmfulness, and alignment-related behavior. \citet{arditi2024refusallanguagemodelsmediated} identify refusal-related directions that can affect model outputs, while subsequent work argues that refusal and safety-related representations need not reduce to a single axis and can involve multiple directions or separable concepts~\citep{pan2025hiddendimensionsllmalignment,wollschlager2025geometryrefusal,zhao2025llmsencodeharmfulnessrefusal}. Closest to our geometry analysis, \citet{shah2025geometryharmfulnesssubconcept} train harmfulness subcategory probes and study the low-rank structure of their probe weights, including steering experiments. Our analysis is narrower: we use difference-of-means contrasts and the row-space of one trained binary final-token probe to characterize where that probe is sensitive or insensitive. We do not interpret these directions as universal harmfulness features or as mechanistic control variables.

\paragraph{Hidden-state trajectories for safety diagnostics.}
Recent work also studies safety- or alignment-relevant information in hidden-state dynamics, including layer-wise evolution, decoding-time trajectories, and latent trajectory classifiers~\citep{zhou2024alignmentjailbreakworkexplain,liu2026trajguardstreaminghiddenstatetrajectory,lin2026nglarenongenerativelatentrepresentationefficient,damirchi2026truthtrajectoryinternalrepresentations}. Our setting is more restricted: within a single prefill pass, we ask whether jailbreak prompts missed by a final-token probe contain earlier probe-visible evidence, and whether a simple trajectory diagnostic trained on the same clean split recovers those misses. We treat the PCA-HMM as a diagnostic complement to the final-token readout, not as a standalone safety monitor.

\section{Experimental Setup}

\paragraph{Models and probes.}
We evaluate three instruction-tuned models: Llama-3.1-8B-Instruct~\citep{llama3.17b}, Mistral-7B-Instruct-v0.1~\citep{jiang2023mistral7b}, and OLMo3-7B-Instruct~\citep{olmo2025olmo3}. For each model, we train SafeSwitch-style probes on final prompt-token hidden states from the original SafeSwitch train split, using clean harmful prompts and benign prompts. Unless otherwise stated, probes are evaluated at threshold 0.5.

\paragraph{Evaluation sets.}
We use held-out SorryBench harmful prompts \citep{xie2024sorrybenchsystematicallyevaluatinglarge} as clean harmful evaluations. To construct semantic jailbreaks, we wrap held-out SorryBench harmful requests with jailbreak templates from HarmBench \citep{mazeika2024harmbenchstandardizedevaluationframework}. We additionally report direct harmful-request evaluations on AdvBench \citep{zou2023universalandtransferableadversarial} and false-positive rates on benign XSTest prompts \citep{rottger2024xstesttestsuiteidentifying}. We report detection rate on harmful and jailbreak sources and false-positive rate on XSTest.

\section{Final-Token Probes}

\subsection{Probe evaluation}

We evaluate the SafeSwitch-style monitoring interface directly: a probe reads the final prompt-token hidden state, projects it through a width-$w$ bottleneck, and predicts unsafe versus safe. We use the released SafeSwitch Llama prober and train analogous Mistral and OLMo3 probers with the same interface, using $w=64$ throughout. We then apply each frozen probe to clean held-out SorryBench prompts, the same SorryBench prompts wrapped with HarmBench jailbreak templates, and benign safety-adjacent XSTest prompts. This tests whether a final-token readout trained on clean harmfulness still exposes the unsafe request after jailbreak wrapping.

\subsection{Probes miss wrapped requests}

Table~\ref{tab:probe-last} reports final-token probe results across the three evaluation conditions. The probes retain high recall on clean held-out SorryBench harmful prompts, but miss a substantial fraction of the same harmful requests once they are wrapped in HarmBench jailbreak templates. The same probes also produce non-trivial false positives on XSTest, a benign set designed to be safety-adjacent. The failure is therefore not a simple thresholding issue: lowering the threshold to recover wrapped harmful requests would further increase false positives on benign safety-adjacent instructions.

\begin{table}[t]
\caption{Final-token probe detection rates on harmful sources and false-positive rate on benign XSTest. Jailbreak prompts use the same held-out SorryBench harmful requests as the clean condition, wrapped with HarmBench templates.}
\label{tab:probe-last}
\centering
\small
\begin{tabular}{lrrr}
\toprule
Model & Sorry & Jailbreak & XSTest \\
\midrule
Llama-3.1-8B & 95.0 & 72.2 & 23.0 \\
Mistral-7B & 94.6 & 86.9 & 32.5 \\
OLMo3-7B & 95.0 & 63.6 & 36.0 \\
\bottomrule
\end{tabular}
\end{table}

This failure has direct downstream consequences. SafeSwitch-style monitoring is a cascade in which the final-token prober gates a stage-2 refusal head: if the prober does not fire on the prefill, the refusal head is never invoked and the wrapped harmful request flows through to generation.

Direct harmful benchmarks rule out a generic distribution-shift explanation. The same final-token probes detect AdvBench at 99.2--99.8\% and HarmBench at 96.2--100.0\% across the three models, so the jailbreak drop is not a generic inability to recognize harmful requests outside the training file; the controlled contrast is clean SorryBench versus the same prompts under adversarial wrapping.

\subsection{Width is not the bottleneck}

SafeSwitch-style probers use a bottlenecked linear readout, so a natural explanation for the wrapped-request failure is that the bottleneck row-space is too narrow to expose the relevant signal. We test this directly by sweeping the bottleneck width from 64 to 1024 with the same training procedure and split.

Wider readouts do not reliably improve jailbreak detection. Going from width 64 to 1024, jailbreak detection moves from 72.2\% to 78.0\% for Llama, from 86.9\% to 87.6\% for Mistral, and from 63.6\% to 55.7\% for OLMo3. For Llama and Mistral the change is small; for OLMo3 the wider probe is worse. The full sweep across widths $\{64, 128, 256, 512, 1024\}$ is reported in Appendix~\ref{app:width-sweep} and shows the same picture. The wrapped-request failure is therefore not just a capacity issue, which motivates looking at where the missed signal sits in representation space rather than how much the probe row-space can hold.

\section{Geometry of Missed Jailbreaks}

Width does not explain the wrapped-request failure, so we ask where the missed signal lives in representation space. Let $h(x) \in \mathbb{R}^D$ denote the model's hidden state at the final prompt token of input $x$. For a finite prompt set $\mathcal{A}$, define
\[
\mu(\mathcal{A}) = \frac{1}{|\mathcal{A}|}\sum_{x\in\mathcal{A}} h(x).
\]
For any nonzero vector $v$, define $u(v)=v/\lVert v\rVert_2$. Let $\mathcal{H}_\mathrm{train}$ and $\mathcal{B}_\mathrm{train}$ be the SafeSwitch-train harmful and benign prompt sets, $\mathcal{S}_\mathrm{sorry}$ be held-out SorryBench harmful prompts, $\mathcal{X}_\mathrm{xstest}$ be benign XSTest prompts, and $\mathcal{J}_\mathrm{caught},\mathcal{J}_\mathrm{miss}$ be the jailbreak prompts caught and missed by the final-token probe. We compute four unit-norm difference-of-means directions:
\begin{align*}
d_\mathrm{harm} &= u\!\left(\mu(\mathcal{H}_\mathrm{train}) - \mu(\mathcal{B}_\mathrm{train})\right), \\
d_\mathrm{safe}  &= u\!\left(\mu(\mathcal{S}_\mathrm{sorry}) - \mu(\mathcal{X}_\mathrm{xstest})\right), \\
d_\mathrm{miss} &= u\!\left(\mu(\mathcal{J}_\mathrm{miss}) - \mu(\mathcal{B}_\mathrm{train})\right), \\
\Delta_\mathrm{cm} &= u\!\left(\mu(\mathcal{J}_\mathrm{caught}) - \mu(\mathcal{J}_\mathrm{miss})\right).
\end{align*}
$d_\mathrm{harm}$ is the clean training-set harmful-vs-benign contrast the prober is trained to find. Motivated by refusal-direction work~\citep{arditi2024refusallanguagemodelsmediated}, $d_\mathrm{safe}$ is a safety-contrast proxy, not the Arditi refusal direction itself: because SorryBench is genuinely harmful and XSTest is safety-adjacent benign, their mean difference contrasts harmful prompts with safety-adjacent benign prompts while partially controlling for refusal-likely surface cues. We keep $d_\mathrm{harm}$ and $d_\mathrm{safe}$ as distinct directions, consistent with recent evidence that LLMs encode harmfulness and refusal as separate internal concepts~\citep{zhao2025llmsencodeharmfulnessrefusal}. $d_\mathrm{miss}$ measures whether missed jailbreaks differ from clean benign at the final token. $\Delta_\mathrm{cm}$ characterizes what separates caught from missed jailbreaks within the jailbreak set.

Let $W \in \mathbb{R}^{w \times D}$ be the prober's first-layer weight matrix. We compute its thin SVD $W = U \Sigma V_r^{\top}$, where $V_r \in \mathbb{R}^{D \times r}$ and $r=\mathrm{rank}(W)\leq w$, and use the right singular vectors as an orthonormal basis for $\mathrm{rowspace}(W)$. Thus $\Pi_W = V_r V_r^{\top}$ is the orthogonal projector onto the probe-visible subspace. For each direction $d \in \mathbb{R}^D$ we report its probe-visible \emph{energy}
\[
\mathrm{E}_W(d) \;=\; \frac{\lVert \Pi_W \, d \rVert_2^{2}}{\lVert d \rVert_2^{2}} \;\in\; [0, 1],
\]
the fraction of $d$'s squared norm captured by the probe row-space. For alignments, we compare the ordinary cosine $\cos(a,b)$ with the probe-visible cosine
\[
\cos_W(a,b) = \frac{\langle \Pi_W a, \Pi_W b\rangle}{\lVert \Pi_W a\rVert_2\lVert \Pi_W b\rVert_2},
\]
using $a=\Delta_\mathrm{cm}$ and $b\in\{d_\mathrm{harm}, d_\mathrm{safe}, d_\mathrm{miss}\}$.

\begin{table}[t]
\caption{Probe-visible energy $\mathrm{E}_W$ of four contrast directions (\%).}
\label{tab:geom-energy}
\centering
\small
\begin{tabular}{lrrr}
\toprule
Direction & Llama & Mistral & OLMo3 \\
\midrule
$d_\mathrm{harm}$ & 9.0 & 12.3 & 12.6 \\
$d_\mathrm{safe}$ & 6.7 & 5.3 & 7.6 \\
$d_\mathrm{miss}$ & 3.5 & 2.4 & 3.2 \\
$\Delta_\mathrm{cm}$ & 4.0 & 5.5 & 4.5 \\
\bottomrule
\end{tabular}
\end{table}

Table~\ref{tab:geom-energy} reports $\mathrm{E}_W$ for each direction. Across all three models, $d_\mathrm{harm}$ has the highest probe-visible energy and $d_\mathrm{miss}$ has the lowest ($3.5\%, 2.4\%, 3.2\%$): the mean direction separating missed jailbreaks from clean benign prompts is the most poorly represented of all four directions in the probe readout, not just at threshold. Equivalently, roughly 96--98\% of $d_\mathrm{miss}$'s squared norm lies in the orthogonal complement of the probe row-space. Moreover, no single SVD basis vector of $W$ aligns with any of the four directions above $|\cos|=0.29$, so even $d_\mathrm{harm}$ is spread thinly across basis vectors rather than captured by one dominant readout direction. This is consistent with harmfulness being a multi-dimensional internal concept~\citep{zhao2025llmsencodeharmfulnessrefusal}.

Full alignment results are reported in Appendix~\ref{app:geom-align}, Table~\ref{tab:geom-align}. Probe projection sharpens alignment with the harmful proxy from $0.32$--$0.39$ to $0.68$--$0.78$, and with the safety proxy from $0.12$--$0.19$ to $0.41$--$0.51$. The full-space anti-alignment with $d_\mathrm{miss}$ is not visible to the probe: after projection, $\Delta_\mathrm{cm}$ has only weak alignment with $d_\mathrm{miss}$ ($0.21$, $0.13$, $0.06$).

Together these findings are consistent with a shortcut-like readout: within the jailbreak distribution, caught prompts are those whose final-token representations retain stronger projection onto the clean harmful-vs-benign direction, while missed prompts move away from this readout. Safety-related cues contribute partially (probe cosine $0.41$--$0.51$ with $d_\mathrm{safe}$), but the strongest projected alignment is with $d_\mathrm{harm}$. Of the four directions tested, the probe row-space is most orthogonal to $d_\mathrm{miss}$, and within probe row-space $\Delta_\mathrm{cm}$ has only weak alignment with $d_\mathrm{miss}$. The wrapper appears to reduce exposure of the relevant feature to the final-token readout, so the probe relies primarily on the clean training-distribution harm proxy.

\section{Unsafe Evidence Appears Before the Last Token}

The geometry above shows that the final-token probe row-space is most blind to the direction separating missed jailbreaks from clean benign prompts. This suggests a positional failure: the unsafe request may be visible earlier in the prompt but no longer exposed at the final token. Recent work studies safety-relevant hidden-state dynamics along other axes, including layer-wise safety evolution, decoding-time token trajectories, multi-probe latent trajectories, and token-layer displacement paths~\citep{zhou2024alignmentjailbreakworkexplain,liu2026trajguardstreaminghiddenstatetrajectory,lin2026nglarenongenerativelatentrepresentationefficient,damirchi2026truthtrajectoryinternalrepresentations}. Our diagnostic is narrower: within a single prefill pass, we sweep prompt-token positions with the same frozen prober and ask whether final-token misses contain earlier probe-visible unsafe evidence.

Across the three models, 87 Llama, 65 Mistral, and 127 OLMo3 final-token-missed jailbreaks had recoverable embedded harmful-request spans. All recoverable cases are detected at the harmful-request span but missed at the final token (Appendix~\ref{app:token-position-results}). Here recoverable means that token matching locates the original harmful request inside the wrapped prompt. Missed jailbreaks are therefore not invisible to the prober everywhere. The failure is positional: by the final token, the wrapped prompt no longer exposes the same probe-visible feature to the prober.

This clarifies the Stage-2 limitation. The same Stage-1 prober can fire on the embedded harmful request when evaluated at the request tokens, but SafeSwitch uses only the final-token Stage-1 decision to gate Stage 2. Thus the downstream refusal head can only refine prompts that pass the final-token gate; it cannot recover probe-visible unsafe evidence that appears earlier and collapses before the readout position.

This observation does not make naive token pooling a usable detector. Across all three models, max-pooling the prober score over prompt positions flags $100\%$ of jailbreak prompts, but it also flags $100\%$ of XSTest prompts. For Llama, XSTest has mean max score $0.998$ despite a final-token score of $0.229$. Thus intermediate high scores are not semantically reliable on their own. The useful object is not whether any token score is high, but how the score evolves across the prompt: missed jailbreaks exhibit a high harmful-request score followed by collapse at the final token. This motivates trajectory-level diagnostics rather than single-position readouts or max-pooling.

\section{Token Trajectories Recover Probe Misses}

The max-pooling result shows that any-position probe scores are too noisy, but the token-position analysis suggests a more structured signal: missed jailbreaks can expose probe-visible unsafe evidence earlier and then collapse at the final token. We therefore use a simple generative sequence diagnostic that asks whether the user-content trajectory is more likely under clean harmful trajectories or clean benign trajectories.

For each model, we fit a 64-dimensional PCA on prompt-token hidden states from balanced clean harmful and benign SafeSwitch training prompts, then fit two 2-state diagonal-covariance Gaussian HMMs in PCA space, one per class. Sequence lengths are passed to the HMM fit. We score a trajectory $z_{1:T}$ by the length-normalized log-likelihood ratio
\[
s(z_{1:T}) =
\frac{1}{T}
\left[
\log p_\mathrm{harm}(z_{1:T})
-
\log p_\mathrm{benign}(z_{1:T})
\right].
\]
The threshold is selected only on the clean training split to maximize harmful-vs-benign training accuracy. We do not tune thresholds on jailbreaks or XSTest.

\begin{figure}[t]
\centering
\includegraphics[width=\columnwidth]{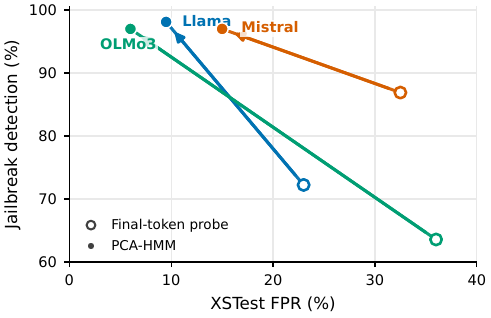}
\caption{Jailbreak--XSTest operating-point shift from the final-token probe to the user-window PCA-HMM trajectory diagnostic. Across all three models, the trajectory diagnostic recovers many final-token misses while reducing XSTest false positives at this operating point.}
\label{fig:jailbreak-xstest-tradeoff}
\end{figure}

\begin{table}[t]
\caption{User-window PCA-HMM diagnostic results. Recovery is measured only among jailbreaks missed by the final-token probe.}
\label{tab:pca-hmm}
\centering
\small
\begin{tabular}{lrrr}
\toprule
Model & Jailbreak & Recovery & XSTest \\
\midrule
Llama-3.1-8B & 98.1 & 94.4 & 9.5 \\
Mistral-7B & 97.0 & 89.8 & 15.0 \\
OLMo3-7B & 97.0 & 94.5 & 6.0 \\
\bottomrule
\end{tabular}
\end{table}

Figure~\ref{fig:jailbreak-xstest-tradeoff} visualizes the operating-point shift, and Table~\ref{tab:pca-hmm} gives the corresponding rates. The diagnostic recovers 236 of 250 Llama misses, 106 of 118 Mistral misses, and 310 of 328 OLMo3 misses without inheriting max-pooling's $100\%$ XSTest false-positive rate.

This result is consistent with the geometry above. The missed-jailbreak direction $d_\mathrm{miss}$ has much higher squared-projection energy in the trajectory PCA subspace than in the final-token probe row-space: $3.5\%\to11.5\%$ for Llama, $2.4\%\to29.2\%$ for Mistral, and $3.2\%\to33.7\%$ for OLMo3. Thus the PCA-HMM is best interpreted as a diagnostic bridge: information that is weakly exposed to the final-token probe can remain recoverable in the prompt trajectory. We do not treat the HMM states as mechanistic features or the diagnostic as a deployed detector.

\section{Limitations and Future Work}

This study is diagnostic rather than a deployed defense. The PCA-HMM threshold is tuned only on clean training data, but false positives remain nonzero and HMM states should not be interpreted as mechanistic safety states. Future work should study calibrated fusion with final-token probes, robustness to adaptive and length-controlled jailbreaks, and causal interventions on trajectory features.

\bibliography{example_paper}

\begin{thebibliography}{17}
\providecommand{\natexlab}[1]{#1}
\providecommand{\url}[1]{\texttt{#1}}
\expandafter\ifx\csname urlstyle\endcsname\relax
  \providecommand{\doi}[1]{doi: #1}\else
  \providecommand{\doi}{doi: \begingroup \urlstyle{rm}\Url}\fi

\bibitem[Arditi et~al.(2024)Arditi, Obeso, Syed, Paleka, Panickssery, Gurnee,
  and Nanda]{arditi2024refusallanguagemodelsmediated}
Arditi, A., Obeso, O., Syed, A., Paleka, D., Panickssery, N., Gurnee, W., and
  Nanda, N.
\newblock Refusal in language models is mediated by a single direction, 2024.
\newblock URL \url{https://arxiv.org/abs/2406.11717}.

\bibitem[Damirchi et~al.(2026)Damirchi, la~Jara, Abbasnejad, Shamsi, Zhang, and
  Shi]{damirchi2026truthtrajectoryinternalrepresentations}
Damirchi, H., la~Jara, I. M.~D., Abbasnejad, E., Shamsi, A., Zhang, Z., and
  Shi, J.
\newblock Truth as a trajectory: What internal representations reveal about
  large language model reasoning, 2026.
\newblock URL \url{https://arxiv.org/abs/2603.01326}.

\bibitem[Han et~al.(2025)Han, Qian, Chen, Zhang, Ji, and
  Zhang]{han2025safeswitchsteeringunsafellm}
Han, P., Qian, C., Chen, X., Zhang, Y., Ji, H., and Zhang, D.
\newblock Safeswitch: Steering unsafe llm behavior via internal activation
  signals, 2025.
\newblock URL \url{https://arxiv.org/abs/2502.01042}.

\bibitem[Jiang et~al.(2023)Jiang, Sablayrolles, Mensch, Bamford, Chaplot,
  de~las Casas, Bressand, Lengyel, Lample, Saulnier, Lavaud, Lachaux, Stock,
  Scao, Lavril, Wang, Lacroix, and Sayed]{jiang2023mistral7b}
Jiang, A.~Q., Sablayrolles, A., Mensch, A., Bamford, C., Chaplot, D.~S., de~las
  Casas, D., Bressand, F., Lengyel, G., Lample, G., Saulnier, L., Lavaud,
  L.~R., Lachaux, M.-A., Stock, P., Scao, T.~L., Lavril, T., Wang, T., Lacroix,
  T., and Sayed, W.~E.
\newblock Mistral 7b, 2023.
\newblock URL \url{https://arxiv.org/abs/2310.06825}.

\bibitem[Lin et~al.(2026)Lin, Yang, Qiu, Guo, Bao, and
  Guan]{lin2026nglarenongenerativelatentrepresentationefficient}
Lin, Z., Yang, J., Qiu, Y., Guo, H., Bao, Y., and Guan, Y.
\newblock N-glare: An non-generative latent representation-efficient llm safety
  evaluator, 2026.
\newblock URL \url{https://arxiv.org/abs/2511.14195}.

\bibitem[Liu et~al.(2026)Liu, Liu, Li, Xin, and
  Ding]{liu2026trajguardstreaminghiddenstatetrajectory}
Liu, C., Liu, X., Li, X., Xin, B., and Ding, K.
\newblock Trajguard: Streaming hidden-state trajectory detection for
  decoding-time jailbreak defense, 2026.
\newblock URL \url{https://arxiv.org/abs/2604.07727}.

\bibitem[Mazeika et~al.(2024)Mazeika, Phan, Yin, Zou, Wang, Mu, Sakhaee, Li,
  Basart, Li, Forsyth, and
  Hendrycks]{mazeika2024harmbenchstandardizedevaluationframework}
Mazeika, M., Phan, L., Yin, X., Zou, A., Wang, Z., Mu, N., Sakhaee, E., Li, N.,
  Basart, S., Li, B., Forsyth, D., and Hendrycks, D.
\newblock Harmbench: A standardized evaluation framework for automated red
  teaming and robust refusal, 2024.
\newblock URL \url{https://arxiv.org/abs/2402.04249}.

\bibitem[Meta(2024)]{llama3.17b}
Meta.
\newblock meta-llama/llama-3.1-8b-instruct, 2024.
\newblock URL \url{https://huggingface.co/meta-llama/Llama-3.1-8B-Instruct}.
\newblock Accessed: 2025-02-21.

\bibitem[Olmo et~al.(2025)Olmo, Ettinger, Bertsch, Kuehl, Graham, Heineman,
  Groeneveld, Brahman, Timbers, Ivison, Morrison, Poznanski, Lo, Soldaini,
  Jordan, Chen, Noukhovitch, Lambert, Walsh, Dasigi, Berry, Malik, Shah, Geng,
  Arora, Gupta, Anderson, Xiao, Murray, Romero, Graf, Asai, Bhagia, Wettig,
  Liu, Rangapur, Anastasiades, Huang, Schwenk, Trivedi, Magnusson, Lochner,
  Liu, Miranda, Sap, Morgan, Schmitz, Guerquin, Wilson, Huff, Bras, Xin, Shao,
  Skjonsberg, Shen, Li, Wilde, Pyatkin, Merrill, Chang, Gu, Zeng, Sabharwal,
  Zettlemoyer, Koh, Farhadi, Smith, and Hajishirzi]{olmo2025olmo3}
Olmo, T., Ettinger, A., Bertsch, A., Kuehl, B., Graham, D., Heineman, D.,
  Groeneveld, D., Brahman, F., Timbers, F., Ivison, H., Morrison, J.,
  Poznanski, J., Lo, K., Soldaini, L., Jordan, M., Chen, M., Noukhovitch, M.,
  Lambert, N., Walsh, P., Dasigi, P., Berry, R., Malik, S., Shah, S., Geng, S.,
  Arora, S., Gupta, S., Anderson, T., Xiao, T., Murray, T., Romero, T., Graf,
  V., Asai, A., Bhagia, A., Wettig, A., Liu, A., Rangapur, A., Anastasiades,
  C., Huang, C., Schwenk, D., Trivedi, H., Magnusson, I., Lochner, J., Liu, J.,
  Miranda, L. J.~V., Sap, M., Morgan, M., Schmitz, M., Guerquin, M., Wilson,
  M., Huff, R., Bras, R.~L., Xin, R., Shao, R., Skjonsberg, S., Shen, S.~Z.,
  Li, S.~S., Wilde, T., Pyatkin, V., Merrill, W., Chang, Y., Gu, Y., Zeng, Z.,
  Sabharwal, A., Zettlemoyer, L., Koh, P.~W., Farhadi, A., Smith, N.~A., and
  Hajishirzi, H.
\newblock Olmo 3, 2025.
\newblock URL \url{https://arxiv.org/abs/2512.13961}.

\bibitem[Pan et~al.(2025)Pan, Liu, Chen, Zhou, Yu, and
  Jia]{pan2025hiddendimensionsllmalignment}
Pan, W., Liu, Z., Chen, Q., Zhou, X., Yu, H., and Jia, X.
\newblock The hidden dimensions of llm alignment: A multi-dimensional analysis
  of orthogonal safety directions, 2025.
\newblock URL \url{https://arxiv.org/abs/2502.09674}.

\bibitem[R{\"o}ttger et~al.(2024)R{\"o}ttger, Kirk, Vidgen, Attanasio, Bianchi,
  and Hovy]{rottger2024xstesttestsuiteidentifying}
R{\"o}ttger, P., Kirk, H.~R., Vidgen, B., Attanasio, G., Bianchi, F., and Hovy,
  D.
\newblock Xstest: A test suite for identifying exaggerated safety behaviours in
  large language models, 2024.
\newblock URL \url{https://arxiv.org/abs/2308.01263}.

\bibitem[Shah et~al.(2025)Shah, Angeline, Kumar, Chheda, Zhu, Sharma, O'Brien,
  and Cai]{shah2025geometryharmfulnesssubconcept}
Shah, M., Angeline, S., Kumar, A.~R., Chheda, N., Zhu, K., Sharma, V., O'Brien,
  S., and Cai, W.
\newblock The geometry of harmfulness in llms through subconcept probing, 2025.
\newblock URL \url{https://arxiv.org/abs/2507.21141}.

\bibitem[Wollschl{\"a}ger et~al.(2025)Wollschl{\"a}ger, Elstner, Geisler,
  Cohen-Addad, G{\"u}nnemann, and Gasteiger]{wollschlager2025geometryrefusal}
Wollschl{\"a}ger, T., Elstner, J., Geisler, S., Cohen-Addad, V., G{\"u}nnemann,
  S., and Gasteiger, J.
\newblock The geometry of refusal in large language models: Concept cones and
  representational independence, 2025.
\newblock URL \url{https://arxiv.org/abs/2502.17420}.

\bibitem[Xie et~al.(2024)Xie, Qi, Zeng, Huang, Sehwag, Huang, He, Wei, Li,
  Sheng, Jia, Li, Li, Chen, Henderson, and
  Mittal]{xie2024sorrybenchsystematicallyevaluatinglarge}
Xie, T., Qi, X., Zeng, Y., Huang, Y., Sehwag, U.~M., Huang, K., He, L., Wei,
  B., Li, D., Sheng, Y., Jia, R., Li, B., Li, K., Chen, D., Henderson, P., and
  Mittal, P.
\newblock Sorry-bench: Systematically evaluating large language model safety
  refusal, 2024.
\newblock URL \url{https://arxiv.org/abs/2406.14598}.

\bibitem[Zhao et~al.(2025)Zhao, Huang, Wu, Bau, and
  Shi]{zhao2025llmsencodeharmfulnessrefusal}
Zhao, J., Huang, J., Wu, Z., Bau, D., and Shi, W.
\newblock Llms encode harmfulness and refusal separately, 2025.
\newblock URL \url{https://arxiv.org/abs/2507.11878}.

\bibitem[Zhou et~al.(2024)Zhou, Yu, Zhang, Xu, Huang, and
  Li]{zhou2024alignmentjailbreakworkexplain}
Zhou, Z., Yu, H., Zhang, X., Xu, R., Huang, F., and Li, Y.
\newblock How alignment and jailbreak work: Explain llm safety through
  intermediate hidden states, 2024.
\newblock URL \url{https://arxiv.org/abs/2406.05644}.

\bibitem[Zou et~al.(2023)Zou, Wang, Carlini, Nasr, Kolter, and
  Fredrikson]{zou2023universalandtransferableadversarial}
Zou, A., Wang, Z., Carlini, N., Nasr, M., Kolter, J.~Z., and Fredrikson, M.
\newblock Universal and transferable adversarial attacks on aligned language
  models, 2023.
\newblock URL \url{https://arxiv.org/abs/2307.15043}.

\end{thebibliography}
\bibliographystyle{icml2026}

\newpage
\appendix
\onecolumn

\section{Bottleneck Width Sweep}
\label{app:width-sweep}

Table~\ref{tab:width-sweep} reports final-token probe jailbreak detection rate as a function of bottleneck width. Wider readouts do not reliably improve detection: Llama gains modestly, Mistral is roughly flat, and OLMo3 degrades.

\begin{table}[h]
\caption{Final-token probe jailbreak detection rate (\%) as a function of bottleneck width.}
\label{tab:width-sweep}
\centering
\small
\begin{tabular}{lrrrrr}
\toprule
Model & 64 & 128 & 256 & 512 & 1024 \\
\midrule
Llama-3.1-8B & 72.2 & 74.3 & 75.8 & 76.3 & 78.0 \\
Mistral-7B & 86.9 & 83.1 & 90.4 & 91.9 & 87.6 \\
OLMo3-7B & 63.6 & 58.2 & 54.6 & 56.2 & 55.7 \\
\bottomrule
\end{tabular}
\end{table}

\section{Geometry Alignment Results}
\label{app:geom-align}

\begin{table}[h]
\caption{Cosine alignment of $\Delta_\mathrm{cm}$ with the harmful, safety-proxy, and missed-jailbreak directions, in the full hidden-state space (Full) and after projection onto the probe row-space (Probe).}
\label{tab:geom-align}
\centering
\small
\begin{tabular}{lrrrrrr}
\toprule
& \multicolumn{2}{c}{Llama} & \multicolumn{2}{c}{Mistral} & \multicolumn{2}{c}{OLMo3} \\
\cmidrule(lr){2-3} \cmidrule(lr){4-5} \cmidrule(lr){6-7}
Comparison & Full & Probe & Full & Probe & Full & Probe \\
\midrule
vs $d_\mathrm{harm}$ & 0.34 & 0.77 & 0.32 & 0.78 & 0.39 & 0.68 \\
vs $d_\mathrm{safe}$ & 0.19 & 0.51 & 0.15 & 0.41 & 0.12 & 0.44 \\
vs $d_\mathrm{miss}$ & -0.26 & 0.21 & -0.30 & 0.13 & -0.05 & 0.06 \\
\bottomrule
\end{tabular}
\end{table}

\section{Trajectory Complementarity}
\label{app:trajectory-complementarity}

\begin{figure}[h]
\centering
\includegraphics[width=0.65\textwidth]{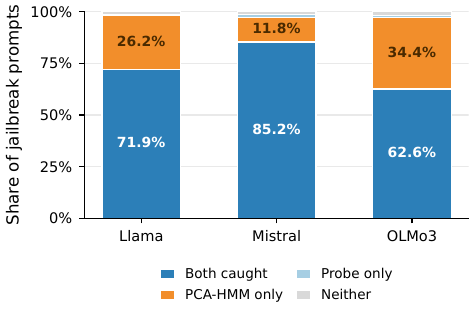}
\caption{Complementarity on jailbreak prompts. Stacked bars partition 900 jailbreak prompts by whether they are caught by the final-token probe, the PCA-HMM trajectory diagnostic, both, or neither. Percentages inside bars are normalized by the 900-prompt jailbreak set. PCA-HMM catches many prompts missed by the final-token probe: 236 for Llama, 106 for Mistral, and 310 for OLMo3.}
\label{fig:jailbreak-complementarity}
\end{figure}

\section{PCA-HMM Length Correlations}
\label{app:length-correlations}

Table~\ref{tab:length-spearman} reports the within-source Spearman rank correlation between user-content window length and PCA-HMM log-likelihood ratio. Correlations are weak to moderate within each source, indicating that the trajectory score is not purely a length proxy within a given evaluation condition. 

\begin{table}[h]
\caption{Within-source Spearman correlation between window length and PCA-HMM score.}
\label{tab:length-spearman}
\centering
\small
\begin{tabular}{llr}
\toprule
Model & Source & Spearman \\
\midrule
Llama-3.1-8B & Jailbreak & 0.29 \\
Llama-3.1-8B & XSTest & 0.31 \\
Mistral-7B & Jailbreak & $-$0.03 \\
Mistral-7B & XSTest & 0.30 \\
OLMo3-7B & Jailbreak & 0.25 \\
OLMo3-7B & XSTest & 0.48 \\
\bottomrule
\end{tabular}
\end{table}

\section{Token-Position Results}
\label{app:token-position-results}

\begin{table}[h]
\caption{Token-position probe scores for final-token-missed jailbreak prompts with located harmful-request spans. Goal is the maximum prober score over the embedded harmful request span.}
\label{tab:token-position-goal}
\centering
\small
\begin{tabular}{lrrrrr}
\toprule
Model & Missed & Located & Located (\%) & Goal & Final \\
\midrule
Llama-3.1-8B & 250 & 87 & 34.8 & 0.998 & 0.174 \\
Mistral-7B & 118 & 65 & 55.1 & 0.998 & 0.168 \\
OLMo3-7B & 328 & 127 & 38.7 & 0.946 & 0.121 \\
\bottomrule
\end{tabular}
\end{table}

All located spans are detected at the harmful-request span and missed at the final token. A span is ``located'' when the original harmful request token sequence can be matched inside the wrapped jailbreak prompt; prompts without a located span are excluded from this span-specific analysis rather than counted as failures.

As a direct-harmful control, Table~\ref{tab:direct-harmful-goal} reports the same span-versus-final-token comparison on sampled clean harmful prompts. Since each direct prompt is itself the harmful request, the request span is located by construction. These prompts remain high at the final token, unlike final-token-missed jailbreaks.

\begin{table}[h]
\caption{Direct harmful prompt control. We sample 50 SorryBench and 50 AdvBench prompts per model. Goal is the maximum prober score over the direct harmful request span.}
\label{tab:direct-harmful-goal}
\centering
\small
\begin{tabular}{llrrrr}
\toprule
Model & Source & $N$ & Goal & Final & Final det. \\
\midrule
Llama-3.1-8B & Sorry & 50 & 0.999 & 0.929 & 94.0\% \\
Llama-3.1-8B & AdvBench & 50 & 1.000 & 0.982 & 98.0\% \\
Mistral-7B & Sorry & 50 & 1.000 & 0.889 & 90.0\% \\
Mistral-7B & AdvBench & 50 & 0.991 & 0.970 & 98.0\% \\
OLMo3-7B & Sorry & 50 & 0.991 & 0.919 & 94.0\% \\
OLMo3-7B & AdvBench & 50 & 0.996 & 0.996 & 100.0\% \\
\bottomrule
\end{tabular}
\end{table}

\end{document}